\documentclass[fleqn,10pt]{wlscirep}
\usepackage{algorithm}
\usepackage{algorithmic}
\usepackage{subfigure}
\usepackage{multirow}
\usepackage{hhline}
\usepackage{amsmath}

\title{Crowdsourcing scoring of immunohistochemistry images: Evaluating Performance of the Crowd and an Automated Computational Method}

\author[1,*]{Humayun Irshad}
\author[1]{Eun-Yeong Oh}
\author[1]{Daniel Schmolze}
\author[1]{Liza M. Quintana}
\author[1]{Laura Collins}
\author[2,3]{Rulla M. Tamimi}
\author[1]{Andrew H. Beck}
\affil[1]{Beth Israel Deaconess Medical Center, Harvard Medical School, Department of Pathology, Boston, 02115, USA}
\affil[2]{Harvard School of Public Health, Department of Epidemiology, Boston, 02115, USA}
\affil[3]{Channing Division of Network Medicine, Brigham and Women's Hospital, Boston, 02115, USA}
\affil[*]{hirshad@bidmc.harvard.edu}


\begin{abstract}
The assessment of protein expression in immunohistochemistry (IHC) images provides important diagnostic, prognostic and predictive information for guiding cancer diagnosis and therapy. Manual scoring of IHC images represents a logistical challenge, as the process is labor intensive and time consuming. Since the last decade, computational methods have been developed to enable the application of quantitative methods for the analysis and interpretation of protein expression in IHC images. These methods have not yet replaced manual scoring for the assessment of IHC in the majority of diagnostic laboratories and in many large-scale research studies. An alternative approach is crowdsourcing the quantification of IHC images to an undefined crowd. The aim of this study is to quantify IHC images for labeling of ER status with two different crowdsourcing approaches, image labeling and nuclei labeling, and compare their performance with automated methods. Crowdsourcing-derived scores obtained greater concordance with the pathologist interpretations for both image labeling and nuclei labeling tasks ($83\%$ and $87\%$), as compared to the pathologist concordance achieved by the automated method ($81\%$) on 5,483 TMA images from 1,909 breast cancer patients.  This analysis shows that crowdsourcing the scoring of protein expression in IHC images is a promising new approach for large scale cancer molecular pathology studies.   
	
\end{abstract}

\begin{document}

\flushbottom
\maketitle
\thispagestyle{empty}

\section*{Introduction}

Immunohistochemistry (IHC) is widely used for measuring the presence and location of protein expression in tissues. The assessment of protein expression by IHC provides important diagnostic, prognostic and predictive information for guiding cancer diagnosis and therapy. In the research setting, IHC is frequently evaluated using tissue microarray (TMA) technology, in which small cores of tissue from hundreds of patients are arrayed on a glass slide, enabling the efficient evaluation of biomarker expression across large numbers of patients. 

The manual pathological scoring of large numbers of TMAs represents a logistical challenge, as the process is labor intensive and time consuming. Over the past decade, computational methods have been developed to enable the application of quantitative methods for the analysis and interpretation of IHC-stained histopathological images \cite{gurcan2009histopathological,irshad2014methods}. While some automated methods have shown high levels of accuracy for IHC markers \cite{giltnane2004technology,bolton2010assessment,ali2013astronomical,howat2015performance}, automated analysis has not yet replaced manual scoring for the assessment of IHC in the majority of diagnostic pathology laboratories and in many large-scale research studies.

In this study, we evaluate the use of crowdsourcing to outsource the task of scoring IHC labeled TMAs to a large crowd of users not previously trained in pathology. Over the last decade, crowdsourcing has been used in a wide range of domains, including astronomy \cite{lintott2008galaxy}, zoology\cite{sullivan2009ebird,marris2010supercomputing,shamir2014classification}, medical microbiology \cite{luengo2012crowdsourcing}, and neuroscience \cite{kim2014space,warby2014sleep,arganda2015crowdsourcing}, to achieve tasks that required large-scale human labeling, which would be difficult or impossible to achieve effectively using only computational methods or domain experts.   

In a  pilot study, we explored the use of crowdsourcing for rapidly obtaining annotations for two core tasks in computational pathology: nucleus detection and segmentation \cite{irshad2015crowdsourcing}. This study concluded that aggregating multiple annotations from a crowd to obtain a consensus annotation could be used effectively to generate large-scale human annotated datasets for nuclei detection and segmentation in histopathological images.  Crowdsourcing has also recently been evaluated for immunohistochemistry studies. Mea et al. crowdsourced 13 IHC images for detection of positive and negative nuclei and reported 0.95 Spearman correlation between pathologist and crowdsourced positivity percentages \cite{della2014preliminary}. Recently, the Cell Slider project \href{http://CellSlider.net}{CellSlider} by Cancer Research UK provided an online interface for members of the general public to score IHC stained TMA images, and they reported high levels of concordance of crowdsourced scores obtained from non-experts and the scores of trained pathologists \cite{dos2015crowdsourcing}.

The purpose of the present study is two-fold. First, we aim to evaluate the performance of crowdsourcing vs. an automated method for scoring protein expression in IHC stained TMA images. Second, we aim to evaluate the time, cost, and accuracy of two different approaches to crowdsourcing the IHC task (image-level labels vs. nucleus-level labels). 

\section*{Methods}

\subsection*{Dataset}

The Nurses’ Health Study (NHS) cohort was established in 1976 when 121,701 female US registered nurses ages 30 to 55 responded to a mail questionnaire that inquired about risk factors for breast cancer \cite{colditz2005nurses}. Every two years, women are sent a questionnaire and asked whether breast cancer has been diagnosed, and if so, the date of diagnosis. All women with reported breast cancers (or the next of kin if deceased) are contacted for permission to review their medical records so as to confirm the diagnosis. Pathology reports are also reviewed to obtain information on ER and PR status. Informed consent was obtained from each participant. This study was approved by the Committee on the Use of Human Subjects in Research at Brigham and Women’s Hospital.

This study used IHC-stained TMA images of breast cancer tissue from the NHS. The dataset consists of $5,483$ scanned images of TMA cores, which were immunostained for estrogen receptor (ER) and scanned using Aperio Slide Scanner at $20\times$ magnification. The average image size is $828 \times 848$ pixels. These images are derived from $1909$ patients, each of whom contributed $1-3$ TMA images, with more than half of the patients contributing $3$ TMA images. All study images were scored by an expert breast pathologist, using three labels (negative=$0$, low positive=$1$ and positive=$2$) \cite{collins2008comparison}. 

\subsection*{Crowdsourcing Platform}
We employed the CrowdFlower platform to design both crowdsourcing applications (image labeling and nuclei labeling). CrowdFlower is a crowdsourcing platform that works with over $50$ labor channel partners to enable access to a network of more than $5$ million contributors worldwide. This platform offers a number of features to improve the likelihood of obtaining high-quality work from contributors. In CrowdFlower, the job designer creates a job in the form of tasks, which are served to contributors for labeling. Each task is a collection of one or more images sampled from the data set.  The job designer creates test questions (test images which have been previously labeled by pathologists) that are used for dual purposes: qualification of contributors during quiz mode and monitoring of contributors during judgment mode. Contributors must maintain a defined level of accuracy on the test questions to be permitted to complete the job. In addition, the job designer specifies the payment per task and the number of labels desired per image. After job completion, CrowdFlower provides a list of labels (annotations) for all the images. Additional information on the CrowdFlower platform is available at www.crowdflower.com. 

\subsection*{Job Design and Crowdsourcing Applications}
Each crowdsourcing job has two modes: quiz mode and judgment mode. Quiz mode occurs at the beginning of a job. In quiz mode, there is only one task and the task consists of $5$ test question images. In judgment mode, there are a number of tasks and each task consists of $4$ actual images and one test image which is presented to the contributor in the same manner as the unlabeled images such that the contributor is unaware if he/she is annotating an unlabeled image or test image. Each contributor must qualify during quiz mode to enter in judgment mode and can remain in judgment mode as long as his/her accuracy on test questions is above a threshold level. For ensuring high quality of labels, we defined five parameters which may influence labeling performance. 

\begin{itemize}
  \item The first is test question minimum accuracy that ensures each contributor must maintain minimum $60\%$ accuracy on test questions throughout the job completion.
  \item The second is minimum time per task that ensures each contributor must spend a minimum of 10 seconds to complete one task.
  \item The third is maximum number of judgments per contributor that enable more contributors to participate in the job. In our jobs, we defined maximum number of judgment per contributor 500 judgments. 
  \item The fourth is a minimum number of images (20) for the contributor to review in work mode prior to computing a trust score for each worker and prior to filtering workers based on their trust score. 
  \item The fifth is the number of labels to collect per image. We collect three labels per image for both jobs. 
\end{itemize}

\begin{figure}[t]
	\centering
	\includegraphics[width=0.7\linewidth]{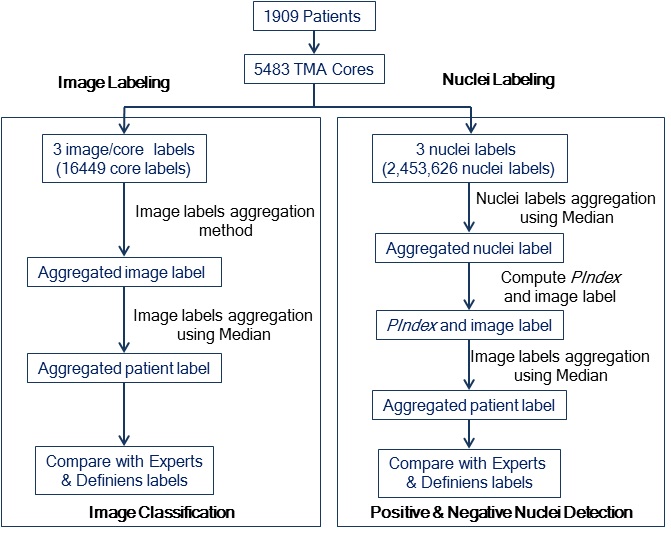}
	\caption{Crowdsourcing work flow for Image Labeling and Nuclei Labeling.}
	\label{fig1:workflow}
\end{figure}

Our study includes two types of labeling jobs: image labeling and nuclei labeling. Figure \ref{fig1:workflow} illustrates the flow chart of both crowdsourcing jobs. Each job contains instructions, which provide examples of expert-derived labels and guidance to assist the contributor in learning the process of labeling.

\subsubsection*{Image Labeling}
\begin{figure}[t]
	\centering
	\includegraphics[width=\linewidth]{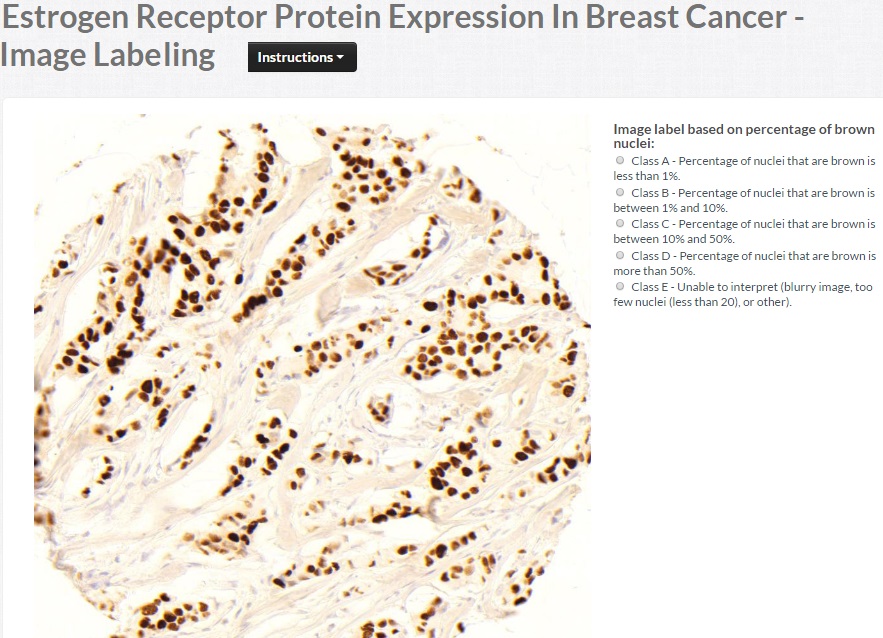}
	\caption{Crowdsourcing Application interface for image labeling. The screenshot illustrates the interface for selecting the image class label.}
	\label{fig2:imagelabel}
\end{figure}

In the image labeling job, each contributor estimates the percentage of cancer nuclei stained brown (positive) and blue (negative) in the image and then selects the image label (score) depending on given criteria: if percentage of brown nuclei is less than $1\%$ then image label is A (negative protein expression), if percentage is between $1\%$ and $10\%$, then image label is B (low positive protein expression), if percentage is between $10\%$ and $50\%$ then image label is C (positive protein expression) and if percentage is more than $50\%$ then image label is D (high positive protein expression). The total pool of test question images used in both quiz and judgment modes are $250$, which are labeled by pathologists. Figure \ref{fig2:imagelabel} shows the interface for image labeling.

\subsubsection*{Nuclei Labeling}
\begin{figure}[ht]
	\centering
	\includegraphics[width=\linewidth]{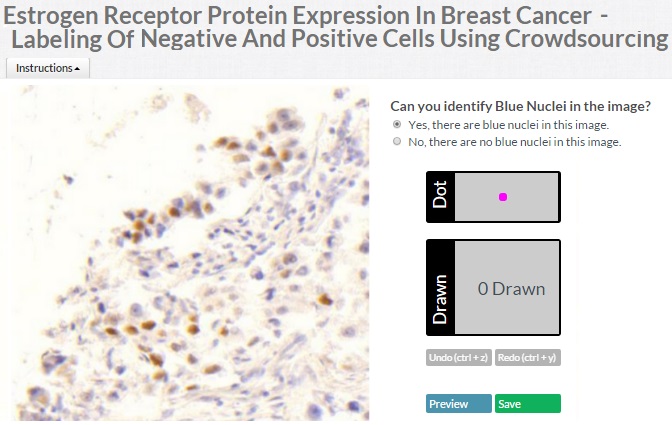}
	\caption{Crowdsourcing application interface for nuclei labeling. The screenshot illustrates the interface for labeling the positive and negative nuclei separately.}
	\label{fig3:nucleilabel}
\end{figure}

In the nuclei labeling job, we ask contributors to detect positive and negative nuclei in the image. In the nuclei labeling job, we first ask contributors to identify the presence of nuclei in the image (yes/no). If they do identify the presence of nuclei, then we ask the contributor to label the nuclei using a dot operator (by clicking at the center of each nucleus). At completion of job, CrowdFlower provides the position of positive and negative nuclei in the images. For each image, we collected positive and negative nuclei from three different contributors. The total pool of test question images used in both quiz and judgment modes for nuclei labeling are $100$ images, which are labeled by pathologists. After counting number of positive and negative nuclei, we compute the positivity index ( $PIndex= \frac{No.of Positive Nuclei}{Total No.of Nuclei}$). From positivity index, we compute the image label using image labeling criteria (mentioned in Image Labeling section). Figure \ref{fig3:nucleilabel} shows the interface for nuclei labeling.

\subsection*{Aggregation Methods for Image Labeling Problem}

We calculated the aggregated label for each image using four different methods: maximum crowd votes (CV), maximum crowd trust scores (CT), maximum weighted crowd votes ($\omega$CV) and maximum weighted crowd trust scores ($\omega$CT). CV is computed by summing the votes for each label and selecting the label with the maximum number of votes as the aggregated label. CT is computed by summing the contributor trust score (CT) for each label and selecting the label with the maximum trust score as the aggregated label. For $\omega$CV and $\omega$CT methods, we multiply the class weights with crowd votes for each label and crowd trust scores for each label, respectively. 

\[
Weighted Crowd Vote= \frac{\omega_A V_A + \omega_B V_B + \omega_C V_C + \omega_D V_D} {V_A + V_B + V_C + V_D} 
\]
\[
Weighted Crowd Trust Score=  \frac{\omega_A T_A + \omega_B T_B + \omega_C T_C + \omega_D T_D} {T_A + T_B + T_C + T_D}
\]

where $V_A$, $V_B$, $V_C$ and $V_D$ are crowd votes from each class labels; $T_A$, $T_B$, $T_C$ and $T_D$ are sum of crowd trust scores for each class labels; and  $\omega_A$, $\omega_B$, $\omega_C$ and $\omega_D$ are class weights. We calculated the class weights by taking the mean of lower and upper boundary of the class. For class A, lower boundary is 0 and upper boundary is $0.01$, the weight of class A is $0.005$. For class B, lower boundary is $0.01$ and upper boundary is $0.1$, the weight of class B is $0.05$. For class C, lower boundary is $0.1$ and upper boundary is $0.5$, the weight of class C is $0.3$. For class D, lower boundary is $0.5$ and upper boundary is $1$, the weight of class D is $0.75$. The aggregated label selected is the label whose class bounds contain the weighted crowd vote or weighted crowd trust score. 

\subsection*{Sensitivity Analysis for Different Combinations of Crowd Size}

To estimate the number of crowd labels required to generate optimal aggregated crowd label, we performed a sensitivity analysis of aggregated labels using different combination of crowd sizes. For this pilot study, we collected 10 crowd labels for each image, and we computed the aggregated label of each image using different combination of crowd sizes (1 to 10), according to Algorithm \ref{algo1}. 

 \begin{algorithm}
 	\begin{algorithmic}
 		\FORALL{ Crowd Sizes: $C_i$ in $i \in 1,2,3, ... , 10$}
 			\STATE $P$ = Compute combination patterns (without replacement) for Crowd Size $C_i$
 			\FORALL {Pattern: $P_j$ in $j \in 1,2,3, ... , J$}
 				\FORALL{ Images: $I_k$ in $k \in 1,2,3, ... , K$}
 					\STATE Compute the Aggregated labels for combination pattern $P_j$ of Crowd Size $C_i$ for Image $I_k$ 
 				\ENDFOR
 			\ENDFOR
 			\STATE Compute Agreement of Aggregated Labels with GT Labels
 		\ENDFOR
 	\end{algorithmic}
 	\caption{Sensitivity Analysis for Different Crowd Sizes}
 	\label{algo1}
 \end{algorithm}

\subsection*{Performance Measures}
We explored different performance measures to evaluate inter-observer reliability of scores. For measuring the inter-observer reliability, we measured percent agreement ($A_g$) or accuracy, which is calculated as the number of agreed labels divided by total number of labels, Kappa ($\kappa$) which measures the agreement among observers adjusted for the possibility of by chance agreement, Spearman correlation ($\rho$) which measures the mean of bivariate Spearman’s rank correlations between observers for inter-observer reliability, and intra-class correlation (ICC).  For image classification, we used $A_g$ performance measures for comparing different methods of label aggregation and the automated method.

\section*{Results}

\subsection*{Image labeling on 380 TMA cores - A pilot study }
We designed a pilot study to test the crowd sourcing application for IHC image labeling and to assess the improvement in crowdsourcing performance as we increase the numbers of aggregated instances per image. In the pilot study, we collected 10 crowdsourced labels for 380 images. We also collected three pathologist labels for each of these 380 images using the same crowd sourcing interface. 

\begin{figure}[t]
	\centering
	\subfigure[Inter-observer reliability of pathologist labels]{
		\includegraphics[width=0.48\linewidth]{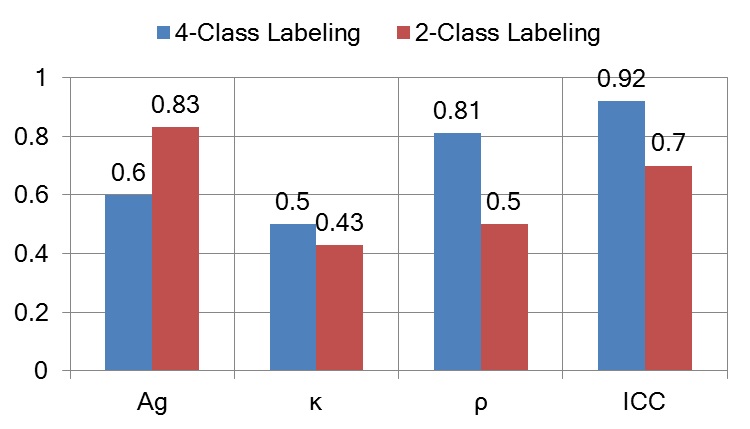}
		\label{fig4a}
	}
	\subfigure[Number of crowd labels agree with pathologist labels]{
		\includegraphics[width=0.48\linewidth]{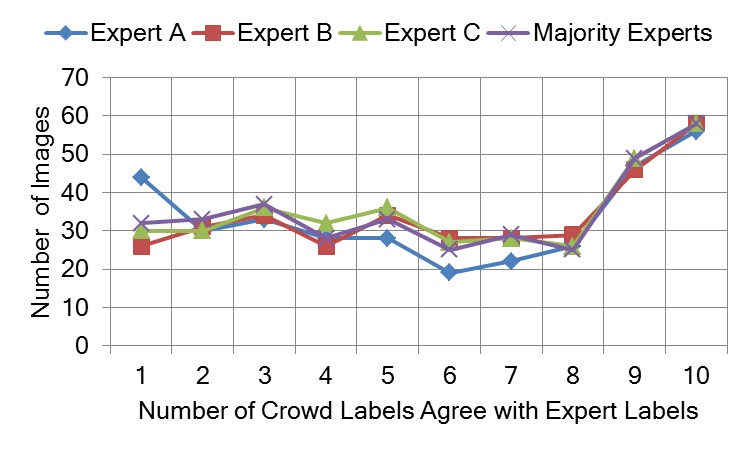}
		\label{fig4b}
	}
	\caption{Inter-observer reliability of pathologist labels and agreement with crowd labels.}
	\label{fig4:expertlabels}
\end{figure}

We assessed inter-observer reliability among pathologists using 4-class labeling as well as 2-class labeling as shown in Figure \ref{fig4a}. For 2-class labeling, we merged all positive classes (B, C and D) into a single positive class (B).  We observed Kappa values of $0.43$ and 0.5 for 2-class and 4-class labeling, respectively, indicating moderate inter-pathologist agreement in IHC interpretation. 

\begin{figure}[t]
	\centering
	\subfigure[4-class labeling sensitivity analysis]{
		\includegraphics[width=0.48\textwidth]{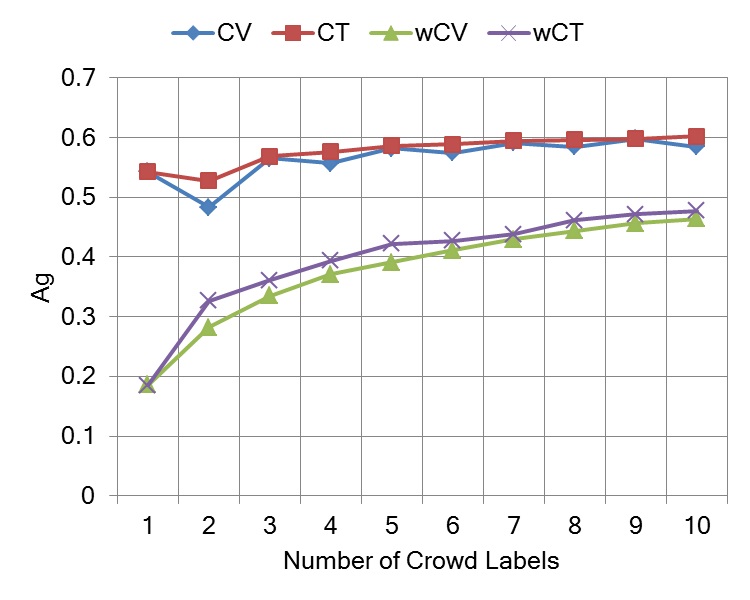}
		\label{fig5a}
	}
	\subfigure[2-class labeling sensitivity analysis]{
		\includegraphics[width=0.48\textwidth]{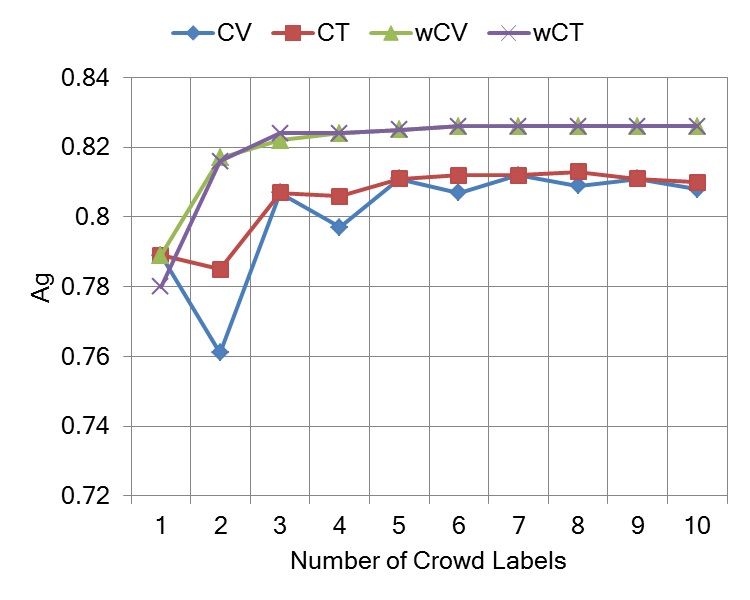}
		\label{fig5b}
	}
	\caption{Sensitivity analysis of crowd labels in pilot study. The analysis supports using 3 crowdsourced labels per image.}
	\label{fig5:sensitivity}
\end{figure}

For 380 images, we obtained 10 crowd labels per image and compared these scores with the pathologist scores as shown in figure \ref{fig4b}. We found a wide range of agreement on images between the crowd and the pathologist scores, with a median level of agreement of 6/10 and $15\%$ of images showing 10/10 agreement.   For a range of crowd labels per image (ranging from 1 to 10), we computed aggregated labels and assessed the agreement of the consensus score with the pathologist score for each number of crowd labels as reported in Figures \ref{fig5a} and \ref{fig5b}. The $A_g$ is not significantly improved after crowd size 3 for 4-class and 2-class image labeling problem.

\subsection*{Image labeling on 5483 TMA cores}

Based on the results of the sensitivity analysis, we collected 3 crowd labels for each image for the image and nucleus labeling crowdsourcing jobs for the main study. The image and nuclei labeling work flow is illustrated in Figure \ref{fig1:workflow}. For 4-class image labeling, we collected 3 labels for each TMA image using CrowdFlower as shown in Figure \ref{fig2:imagelabel}. In total, 16,449 image labels were collected for 5,483 images. Aggregated image labels were computed with four aggregation methods (CV, CT, $\omega$CV and $\omega$CT). The aggregated label at patient level was computed by taking the median of all aggregated image labels belonging to that patient. Pathologists labeled these images using 3-class labeling: negative (A), low positive (B) and positive (C). Since we obtained 4-class labeling from the crowd, to compare the crowd labels with pathologist labels, we merged crowd class D into crowd class C. The CV aggregation method reported higher $A_g$ and Spearman $\rho$ than other aggregation methods for 3-class labeling as reported in Table \ref{Tab1}. For 2-class labeling, we merged all positive classes (B and C) into a single positive class (B) for both crowd and pathologist aggregated labels. The CV aggregation method outperformed as compared to other aggregation method for $A_g$ and $\rho$ as reported in Table \ref{Tab1}. 

\begin{table}[t]
  \centering
  \begin{tabular}{|c|c|c|c|c|c|}
    \hline
    \multirow{2}{*}{Crowdsourcing Types} & \multirow{2}{*}{Methos} & \multicolumn{2}{|c|}{3-Class Labeling} & \multicolumn{2}{|c|}{2-Class Labeling} \\
    \hhline{~~----}
    &  & $A_g$ & $\rho$ & $A_g$ & $\rho$ \\
    \hline
    \multirow{2}{*}{Image Labeling} & Crowd CV & \textbf{0.71} & 0.64 & \textbf{0.83} & \textbf{0.62} \\
    \hhline{~-----}
	 & Crowd CT & 0.68 & \textbf{0.65} & 0.81 & 0.61 \\    
    \hhline{~-----}
     & Crowd $\omega$CV & 0.64 & 0.63 & 0.77 & 0.59 \\
    \hhline{~-----}
     & Crowd $\omega$CT & 0.64 & 0.64 & 0.77 & 0.59 \\
    \hline
    \multirow{2}{*}{Nuclei Labeling} & Crowd & \textbf{0.77} & \textbf{0.68} & \textbf{0.87} & \textbf{0.63} \\
    \hhline{~-----}
	 & Definiens & 0.70 & 0.51 & 0.81 & 0.48 \\
	\hline
  \end{tabular}
  \caption{Crowdsourced image labeling and nuclei labeling results.}
  \label{Tab1}
\end{table}

\subsection*{Nuclei Labeling on 5483 TMA cores}

For the nuclei labeling job, we collected 3 nuclei labels for all 5,483 TMA images. Total number of nuclei labels was 2,453,646. The aggregated number of positive and negative nuclei was calculated for each image as the median number of positive and negative nuclei labeled by the crowd. Then, we computed $PIndex$ for each image. $PIndex$ was converted into image labels (A,B and C) according to following image labeling criteria:

\begin{equation*}
	\text{Image Label} = 
	\begin{cases}
	 \text{Class C (positive)}     & \quad \text{if } \quad PIndex > 0.1 \\
	 \text{Class B (low positive)} & \quad \text{if } \quad PIndex > 0.01 \quad \text{and} \quad PIndex \leq 0.1 \\
	 \text{Class A (negative)}     & \quad \text{if } \quad PIndex \leq 0.01
	\end{cases}
\end{equation*}

Lastly, we computed the aggregated patient label by taking the median of the all the image labels belonging to that patient and compared with the pathologist labels as reported in Table \ref{Tab1}. We also performed 2-class labeling by merging all positive classes into a single positive class for both crowd and pathologist labels. The $A_g$ and $\rho$ are 0.77 and 0.68 for 3-class labeling and 0.87 and 0.63 for 2-class labeling, respectively.

In order to compare with an automated method, we developed an image processing pipeline in Definiens Tissue Studio. This pipeline detected positive and negative nuclei in TMA images and computed the $PIndex$. The crowdsourcing $PIndex$ was correlated with the Definiens $PIndex$ ($\rho$ is 0.75). However, considering pathologist labels as ground truth, both types of crowdsourcing jobs (image and nuclei labeling) resulted in higher $A_g$ and $\rho$ than Definiens for both 3-class labeling and 2-class labeling. 

\begin{table}[b]
  \centering
  \begin{tabular}{|c|c|c|c|c|}
    \hline
    \multirow{2}{*}{Crowdsourcing Jobs} & \multicolumn{2}{|c|}{Quiz Mode} & \multicolumn{2}{|c|}{Work Mode} \\
    \hhline{~----}
    &  Passed & Failed & Passed & Failed \\
    \hline
	Image Labeling & 113 & 155 & 61 & 52 \\
    \hhline{-----}
    Nuclei Labeling & 3244 & 1572 & 2216 & 1243 \\
	\hline
  \end{tabular}
  \caption{Crowd performance on test questions in quiz mode and work mode.}
  \label{Tab2}
\end{table}

The Crowd showed significantly improved performance on test questions for the nuclei labeling job as compared with the image labeling task as reported in Table \ref{Tab2}. This finding supports the overall higher level of accuracy seen with the nuclei labeling approach as compared with the image labeling approach. 

\subsection*{Crowdsourcing Performance}

\begin{figure}[t]
	\centering
	\subfigure[Contributor trust score]{
		\includegraphics[width=0.32\textwidth]{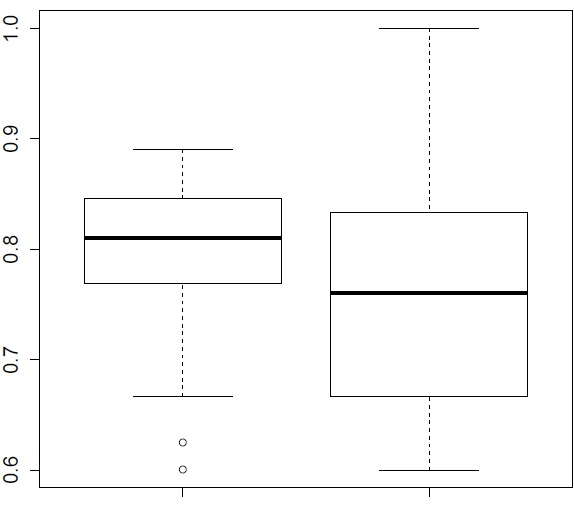}
		\label{fig6a}
	}
	\subfigure[Contributor trust scores on image labeling ]{
		\includegraphics[width=0.32\textwidth]{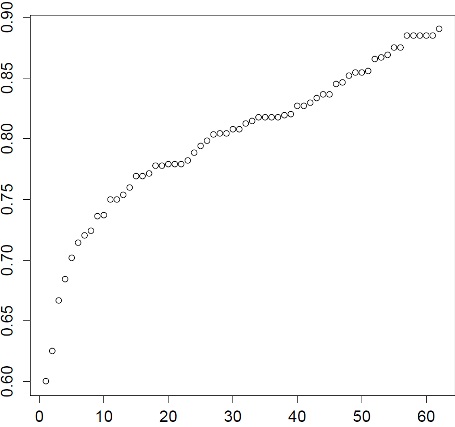}
		\label{fig6b}
	}
	\subfigure[Contributor trust scores on nuclei labeling]{
		\includegraphics[width=0.32\textwidth]{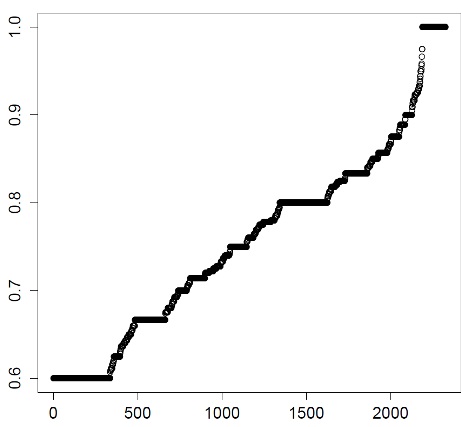}
		\label{fig6c}
	}
	\caption{Contributor trust scores analysis on Crowdsourcing jobs.}
	\label{fig6}
\end{figure}

We first assessed the contributor (crowd) performance for both crowdsourcing jobs. The number of contributors who participated in both jobs is shown in Table \ref{Tab2}. The contributors who maintained the minimum accuracy $(60\%)$ on test questions during quiz and work modes are trusted contributors and the rest are untrusted contributors. In work mode, there were 61 trusted contributors for image labeling and 2,216 for nuclei labeling. The average time of trusted contributors was 32 seconds for image labeling and 306 seconds for nuclei labeling per image while the average time of untrusted contributors was 149 seconds for image labeling and 207 seconds for nuclei labeling. Thus, trusted contributors took less time to label images as compared to untrusted contributors; however, trusted contributors took more time to label nuclei as compared to untrusted contributors. These results suggest that efficient labeling of nuclei is a complex job requiring sufficient time for strong performance. Figure \ref{fig6} illustrates the distribution of crowd trust scores for both jobs. The image labeling contributors have higher trust score as compared to nuclei labeling contributors. The average test question accuracy for trusted contributors is $80\%$ for image labeling and $76\%$ for nuclei labeling while average test question accuracy for untrusted contributors is $66\%$ for image labeling and $42\%$ for nuclei labeling. The trust scores were moderately correlated with the number of images labeled; $\rho=0.41$, $P < 0.0008$ for image labeling job and $\rho=0.186$, $P < 2.2e^{-16}$ for nuclei labeling. The average time for image labeling is 50 seconds and nuclei labeling per image is 373 seconds. 

The image labeling job was finished in 4 hours and nuclei labeling was finished in 472 hours. The Crowdflower platform charged \$ 282 for image labeling and \$ 2,280 for nuclei labeling job. These data suggest that although nuclei labeling produced some improvements in accuracy, it cost significantly longer in terms of time to complete the full job (118 fold longer) and cost (~8 fold more expensive).

\section*{Discussion}

The principle of applying crowdsourcing in science, which has enabled whale sound classification \cite{shamir2014classification}, malaria parasite classification \cite{luengo2012crowdsourcing}, sleep spindle detection \cite{warby2014sleep} and nuclei detection and segmentation in histopathology \cite{irshad2015crowdsourcing}, has become increasingly well established in recent years. Crowdsourced work can be used to classify objects (whale sound and malaria parasite classification), detect objects (nuclei detection) and segment objects (nuclei segmentation).  The aim of this study was to better understand how to use crowdsourcing for IHC image interpretation. This laborious and time consuming image quantification task has also been performed using automated methods \cite{bolton2010assessment, ali2013astronomical, howat2015performance, mohammed2012comparison, inwald2013ki,gudlaugsson2012comparison}. However, no prior studies have directly compared crowdsourcing vs. automated methods in the interpretation of IHC.

In this study, we quantify IHC TMA images for labeling of ER status with two different crowdsourcing approaches, image labeling and nuclei labeling. In the image labeling task, the crowd was asked to estimate the percentage of positive cells for each IHC image, while in nuclei labeling task, the crowd was asked to label individual nuclei within IHC images as either positive or negative. We completed these crowdsourcing tasks on a large data set containing 5483 TMA images belonging to 1909 patients, which were previously labeled by an expert pathologist and by an automated method. In our study, crowdsourcing-derived scores obtained greater concordance with the pathologist interpretation for both image labeling and nuclei labeling tasks, as compared with the pathologist concordance achieved by the automated method.
 
Overall, the crowdsourced scores produced from nuclei labeling (as opposed to image labeling) showed somewhat higher agreement with the pathologist scores; however, the time and cost required for the nuclei labeling far exceeded the time and cost for the image labeling. 
Nuclei labeling is more laborious task spread over many people, even though paying more and takes longer time for nuclei scoring, still cheaper and quicker than using pathologists. Our study results support that crowdsourcing is a promising new approach for scoring biomarker studies in large scale cancer molecular pathology studies. A limitation of our current crowdsourcing application is that we do not ask the Crowd to classify nuclei into specific types (e.g., cancer epithelial nucleus, lymphocyte nucleus). We expect the addition of training the crowd to classify cell types in addition to classifying IHC positivity will further improve crowd performance, although the incorporation of cell type-specific scoring may increase the time and cost of the overall task. This represents an important direction for future research.   

\bibliography{sample}

\section*{Acknowledgements}

The research reported in this publication was supported in part by the National Library of Medicine of the National Institutes of Health under Award Number K22LM011931. We would like to thank the participants and staff of the Nurses' Health Study (NHS) for their valuable contributions as well as the following state cancer registries for their help: AL, AZ, AR, CA, CO, CT, DE, FL, GA, ID, IL, IN, IA, KY, LA, ME, MD, MA, MI, NE, NH, NJ, NY, NC, ND, OH, OK, OR, PA, RI, SC, TN, TX, VA, WA, WY. The authors assume full responsibility for analyses and interpretation of these data. The data collection and creation of TMAs in this publication was supported by Public Health Service Grant under Award Number CA087969 and National Cancer Institute of National Institutes of Health under Award Number CA186107. This investigation was approved by the institutional Review Board at the Brigham and Women's Hospital and the Harvard School of Public Health. 

\section*{Author contributions statement}

H.I. and A.H.B. designed the study. H.I. developed crowdsourcing application, performed experiments and analyzed the results. E.O., D.S., L.M.Q and L.C. are pathologists and performed image labeling. R.M.T. provides NHS data. All authors reviewed the manuscript. 

\section*{Competing Financial Interests}

AHB has an equity interest in PathAI, Inc.

\end{document}